\documentclass[runningheads]{llncs}
\usepackage{graphicx}
\usepackage{amsmath,amssymb} % define this before the line numbering.
\usepackage{ruler}
\usepackage{color}
\usepackage[width=122mm,left=12mm,paperwidth=146mm,height=193mm,top=12mm,paperheight=217mm]{geometry}
\usepackage{times}
\usepackage{epsfig}
\usepackage{array}
\usepackage{subcaption}
\usepackage{caption}
\usepackage{booktabs}
\usepackage{siunitx}
\usepackage{multirow}
\usepackage{tabularx}
\usepackage{float}
\usepackage{xcolor}
\usepackage{multicol}

\graphicspath{ {figback/} }
\usepackage[pagebackref=true,breaklinks=true,letterpaper=true,colorlinks,bookmarks=false]{hyperref}

\makeatletter
\def\fixedlabel#1#2{%
  \@bsphack
  \begingroup
    \@onelevel@sanitize\@currentlabelname
    \edef\@currentlabelname{%
      \expandafter\strip@period\@currentlabelname\relax.\relax\@@@%
    }%
    \phantomsection%
    \protected@write\@auxout{}{%
      \string\newlabel{#1}{%
        {#2}%
        {\thepage}%
        {#2}%
        {\@currentHref}{}%
      }%
    }%
  \endgroup
  \@esphack
}
\makeatother

\newcommand{\specialcell}[2][c]{%
  \begin{tabular}[#1]{@{}c@{}}#2\end{tabular}}

\newcommand\crule[3][black]{\textcolor{#1}{\rule{#2}{#3}}}

\newcommand{\etal}{\textit{et al}.}

\begin{document}
% \linenumbers
\pagestyle{headings}
\mainmatter

\title{Design of Kernels in Convolutional Neural Networks for Image Classification} % Replace with your title

% \titlerunning{ECCV-16 submission ID \ECCV16SubNumber}

% \authorrunning{ECCV-16 submission ID \ECCV16SubNumber}

\author{Zhun Sun \ \ \ \ Mete Ozay \ \ \ \ Takayuki Okatani\\}
\institute{\{sun, mozay, okatani\}@vision.is.tohoku.ac.jp}

\maketitle

\begin{abstract}
Despite the effectiveness of convolutional neural networks (CNNs) for image classification, our understanding of the effect of shape of convolution kernels on learned representations is limited. In this work, we explore and employ the relationship between shape of kernels which define receptive fields (RFs) in CNNs for learning of feature representations and image classification. For this purpose, we present a feature visualization method for visualization of pixel-wise classification score maps of learned features. Motivated by our experimental results, and observations reported in the literature for modeling of visual systems, we propose a novel design of shape of kernels for learning of representations in CNNs. 

In the experimental results, the proposed models also outperform the state-of-the-art methods employed on the CIFAR-10/100 datasets \cite{cifar} for image classification. We also achieved an outstanding performance in the  classification task, comparing to a base CNN model that introduces more parameters and computational time, using the ILSVRC-2012 dataset~\cite{ILSVRC15}. Additionally, we examined the region of interest (ROI) of different models in the classification task and analyzed the robustness of the proposed method to occluded images. Our results indicate the effectiveness of the proposed approach.

\end{abstract}

\section{Introduction}
\label{sec:intro}

Following the success of convolutional neural networks (CNNs) for large scale image classification \cite{ILSVRC15,Alexnet}, remarkable efforts have been made to deliver state-of-the-art performance on this task. Along with more complex and elaborate architectures, lots of techniques concerning parameter initialization, optimization and regularization have also been developed to achieve better performance. Despite the fact that various aspects of CNNs have been investigated, design of the convolution kernels, which can be considered as one of the fundamental problems, has been barely studied. Some studies examined how size of kernels affects performance \cite{VGG}, leading to a recent trend of stacking small kernels (e.g. $3\times3$) in deep layers of CNNs. However, analysis of the shapes of kernels is mostly left untouched. Although there seems to be no latitude in designing the shape of convolution kernels intuitively (especially ${3\times 3}$ kernels), in this work, we suggest that designing the shapes of kernels is feasible and practical, and we analyze its effect on the performance.

% Assuming the use of ${3\times 3}$ kernels, there does not seem to be much room for designing their shapes. In this paper we show that, contrary to this intuition, designing the shapes of kernels is indeed possible and practically useful.

%The only exceptions are a few studies aiming at learning shapes of the receptive fields (RFs) of units in intermediate layers by selecting from a set of overcomplete RFs \cite{selectRF,JiaRF,CollaRF}. 

In the early studies of biological vision \cite{hw,lowe,simon}, it was observed that the receptive fields (RFs) of neurons are arranged in an approximately hexagonal lattice. A recent work reported an interesting result that an irregular lattice with appropriately adjusted asymmetric RFs can be accurate in representation of visual patterns \cite{pred}. Intriguingly, hexagonal-shaped filters and lattice structures have been analyzed and employed for solving various problems in computer vision and image processing \cite{hexsnn,1979}. In this work, motivated by these studies, we propose a method for designing the kernel shapes in CNNs. Specifically, we propose a method to use an asymmetric shape, which simulates hexagonal lattices, for convolution kernels (see Fig.~\ref{fig3} and \ref{fig4}), and then deploy kernels with this shape in different orientations for different layers of CNNs (Sect.~\ref{sec:method}).

\begin{figure}[tH]
\centering
    %for row 1
    \begin{minipage}[c]{.32\linewidth}
    \centering
    \includegraphics[width=1\textwidth]{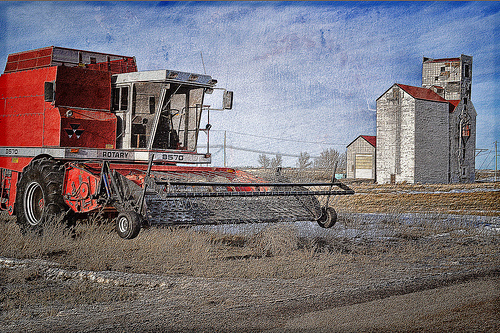}
    \end{minipage}
    \begin{minipage}[c]{.32\linewidth}
    \centering
    \includegraphics[width=1\textwidth]{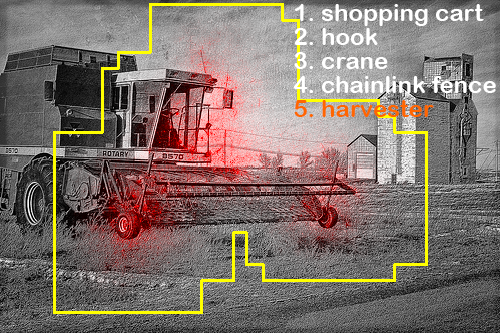}
    \end{minipage}
    \begin{minipage}[c]{.32\linewidth}
    \centering
    \includegraphics[width=1\textwidth]{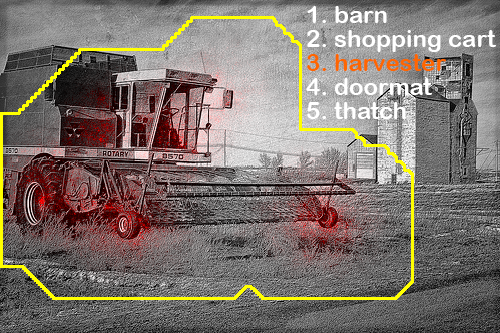}
    \end{minipage}
    
    %for row 2
    \begin{minipage}[c]{.32\linewidth}
    \centering
    \includegraphics[width=1\textwidth]{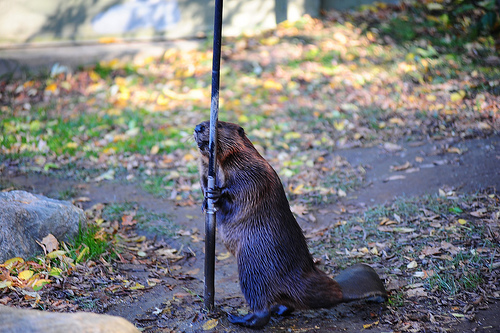}
    \end{minipage}
    \begin{minipage}[c]{.32\linewidth}
    \centering
    \includegraphics[width=1\textwidth]{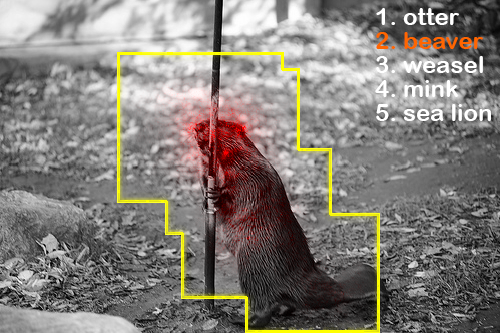}
    \end{minipage}
    \begin{minipage}[c]{.32\linewidth}
    \centering
    \includegraphics[width=1\textwidth]{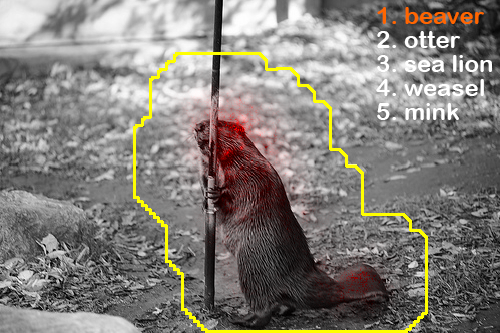}
    \end{minipage}

    %for row 3
    \begin{minipage}[c]{.32\linewidth}
    \centering
    \subcaption{}%Original images belonging to harvester and beaver classes.}
    \end{minipage}
    \begin{minipage}[c]{.32\linewidth}
    \centering
    \subcaption{}%ROI obtained using square kernel.}
    \label{fig1b}
    \end{minipage}
    \begin{minipage}[c]{.32\linewidth}
    \centering
    \subcaption{}%ROI obtained using the proposed ``quasi-hexagonal'' kernel.}
    \label{fig1c}
    \end{minipage}
    
\vspace{-0.1in}
\caption{Examples of visualization of ROI (Sect.~\ref{sec:vis}) in two images (a) for CNNs equipped with kernels with (b) square, and (c) our proposed ``quasi-hexagonal'' shapes (Sect.~\ref{sec:method}). The pixels marked with red color indicate their maximum contribution for classification scores of the correct classes. For (b), these pixels tend to be concentrated on local, specific parts of the object, whereas for (c), they distribute more across multiple local parts of the object. See texts for more details.}
%models. Detected features located at the pixels are covered by image regions which are defined by RFs. Boundaries of the regions are depicted using yellow frames. The convolution operation implemented using kernels with square shape may accumulate features on more confined regions to represent discriminative patterns (b). RFs can be employed for representation of spatially distributed patterns, and for improvement of class predictions using our proposed method (c).}
\vspace{-0.1in}
%In the CNN models that employ square shape kernels, which would lead two mean issues: unnatural expression of objects with rigid-shape RFs, and the non-spatially distributed representations, as shown in the second row. We attempted to address these issues with a proposed topology guided architecture, by which we could obtain appropriate shaped RFs and more spatially distributed representations as shown in the last row.}
\label{fig1}
\end{figure}

This design of kernel shapes brings multiple advantages. Firstly, as will be shown in the experimental results (Sect.~\ref{sec:perf}), CNNs which employ the proposed design method are able to achieve comparable or even better classification performance, compared to CNNs which are constructed using the same architectures (same depth and output channels for each layer) but employing square (${3\times 3}$) kernels. Thus, a notable improvement in computational efficiency (a reduction of 22\% parameters and training time) can be achieved as the proposed kernels include  fewer weights than ${3\times 3}$ kernels. Meanwhile, increasing the number of output channels of our proposed models (to keep the number of parameters same as corresponding models with square shape), leads to a further improvement in performance.

% Secondly, as will also be shown in the experimental results, the CNNs with our kernels show higher performance than those with the square kernels, when increasing the number of kernels by about 22\% for the former so that they have the same number of parameters as the latter. This is not an obvious result, because this occurs when the CNNs with the square kernels are well optimized and thus simply increasing the number of kernels in them does not improve their performance. It is theoretically possible that the $3\times3$ square kernels learn, if necessary, exactly the same shapes of our asymmetric kernels, since our kernel design is equivalent to setting the chosen weights of the square kernel to zero. The experimental results imply that our kernel design also serves as a kind of regularization in training. 

Secondly, CNNs which employ our proposed kernels provide improvement in learning for extraction of discriminative features in a more flexible and robust manner. This results in better robustness to various types of noise in natural images that could make classification erroneous, such as occlusions. Fig.~\ref{fig1} shows examples of visualization of features extracted  using fully-trained CNNs equipped with and without our proposed kernels, which are obtained by the method introduced in Sec. \ref{sec:vis}. These depict the image pixels that have the maximum contribution to the classification score of the correct class (shown in red). It is observed that for CNNs equipped with our proposed kernels, they tend to be less concentrated on local regions and rather distributed across a number of sub-regions, as compared to CNNs  with standard square kernels. This property prevents erroneous classification due to occlusions, as will be shown in the experimental results. This also helps to explain the fact that the CNNs equipped with our proposed kernels perform on par with the CNNs equipped with square kernels despite having less number of parameters.
% ; and that the former outperforms the latter when their numbers of parameters are equal. 

The contributions of the paper are summarized as follows:
\begin{enumerate}
\item We propose a method to design  convolution kernels in deep layers of CNNs, which is inspired by hexagonal lattice structures employed for solving various problems of computer vision and image processing.
\item We examine  classification performance of  CNNs equipped with our kernels, and compare the results with state-of-the-art CNNs equipped with square kernels using benchmark datasets, namely ImageNet and CIFAR 10/100. The experimental results show that the proposed method is superior to the state-of-the-art CNN models in terms of computational time and/or classification performance.
\item We introduce a method for visualization of features to qualitatively analyze the effect of kernel design on classification. Additionally, we analyze the robustness of CNNs equipped with and without our kernel design to occlusion  by measuring their classification accuracy when some regions on input images are occluded.
%, where in the input image CNNs are looking at to perform classification, using which we analyze the effect of kernel design on classification. 

\end{enumerate}
\vspace{-0.1in}

\begin{figure}[tH]
\centering
    %for row 1
    \begin{minipage}{0.42\linewidth}
    \includegraphics[width=1\textwidth]{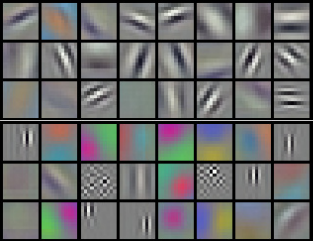}
    \subcaption{$ \{ \mathbf{K}^S_{a,l=1} \in {\mathbb{R}^{K \times K}} \} _{a=1} ^{48}$.}\label{fig2a}
    \end{minipage}
    \begin{minipage}{0.1\linewidth}
    \crule[white]{10pt}{1pt}
    \end{minipage}
    \begin{minipage}{0.42\linewidth}
    \includegraphics[width=1\textwidth]{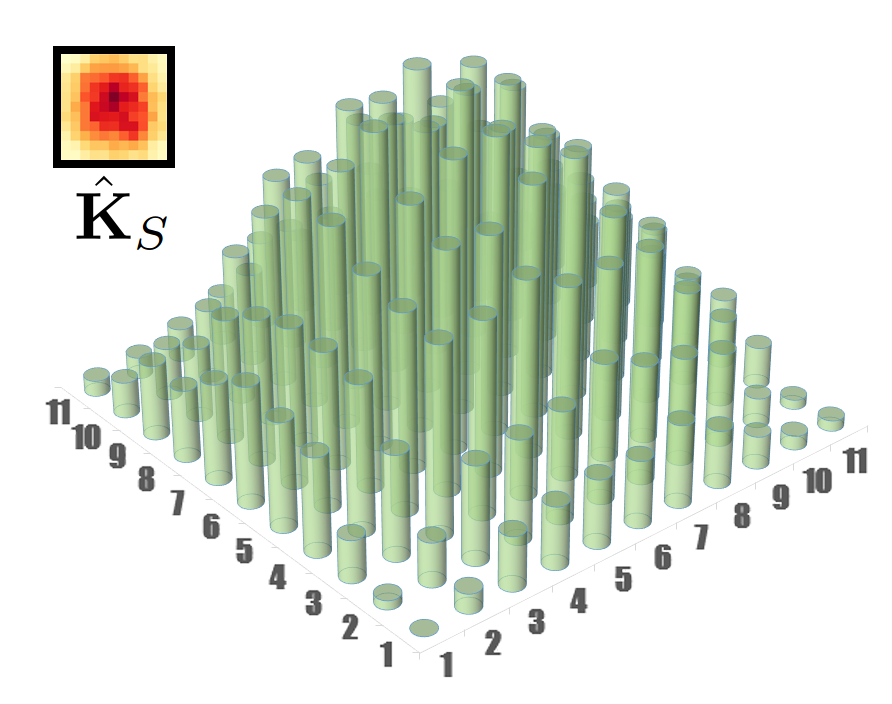}
    \vspace{-0.25in}
    \subcaption{$\frac{1}{A}\sum_{c}|\mathbf{K}^S_{a,1}(i,j,c)|, \forall i,j,a$}\label{fig2b}
    \end{minipage}
\caption{(a) Visualization of a subset of kernels $\mathbf{K}^S_{a,l} \in {\mathbb{R}^{K \times K}}$, where $K$ is the size of kernel, at the first convolution layer $l=1$ of AlexNet \cite{Alexnet} trained on ImageNet. (b) An average kernel $\hat{\mathbf{K}}_S=\frac{1}{A} \sum_{a=1}^{A}|\mathbf{K}^S_{a,l}|$ is depicted at the top-left part. Each bar in the histogram shows a cumulative distribution of values over each channel, $c$.}
\vspace{-0.1in}
\label{fig2}
\end{figure}

%------------------------------------------------------------------------
\section{Our approach}
\label{sec:method}

We propose a method for designing shape of convolution kernels which will be employed for image classification. The proposed method enables us to reduce the computational time of training CNNs providing more compact representations, while preserving the classification performance.

In CNNs~\cite{Alexnet,LeNet,VGG}, an input image (or feature map) $\mathbf{I}\in{\mathbb{R}^{W\times H \times C}}$ is convolved with a series of square shaped kernels $\mathbf{K}^S\in{\mathbb{R}^{K\times K \times C}}$ through its hierarchy. The convolution operation $\mathbf{K}^S*\mathbf{I}$ can be considered as sampling of the image $\mathbf{I}$, and extraction of discriminative information with learned representations.
Fig.~\ref{fig2} shows a subset of learned kernels $\mathbf{K}^S$, and the kernel $\hat{\mathbf{K}}_S$ averaged over all the kernels employed at the first layer of AlexNet~\cite{Alexnet}. Distribution of values of $\hat{\mathbf{K}}_S$ shows that most of the weights at the corner take values close to zero, thus making less contribution for representing features at the higher layers. If a computationally efficient and compressed model is desired, additional methods need to be employed, such as pruning these diluted parameters during fine-tuning~\cite{Learnweight}.

\begin{figure}[t]
\centering
    %%%%%%%%%%%%%%%%%%% ROW 1 %%%%%%%%%%%%%%%%%%%%
    \begin{minipage}{0.32\linewidth}
    \includegraphics[width=1\textwidth]{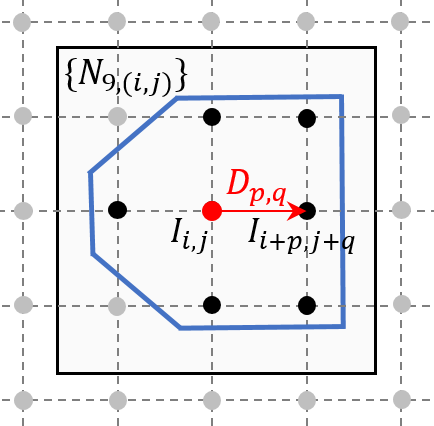}
    \end{minipage}
    \begin{minipage}{0.32\linewidth}
    \includegraphics[width=1\textwidth]{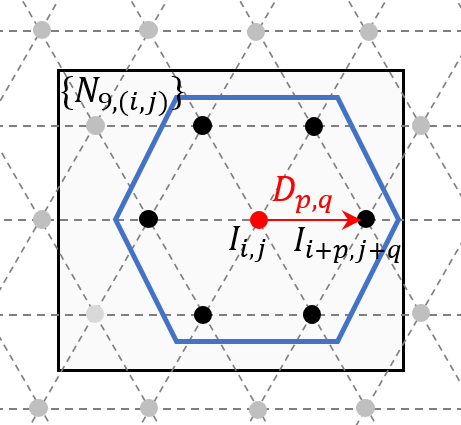}
    \end{minipage}
    \begin{minipage}{0.32\linewidth}
    \begin{minipage}{0.49\linewidth}
    \includegraphics[width=1\textwidth]{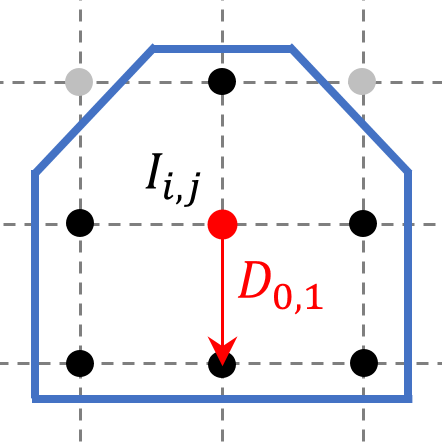}
    \end{minipage}
    \begin{minipage}{0.49\linewidth}
    \includegraphics[width=1\textwidth]{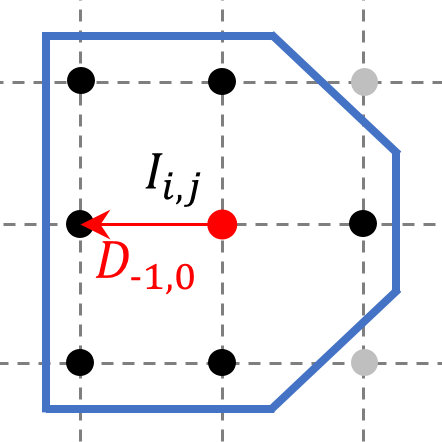}
    \end{minipage}
    
    \begin{minipage}{0.49\linewidth}
    \includegraphics[width=1\textwidth]{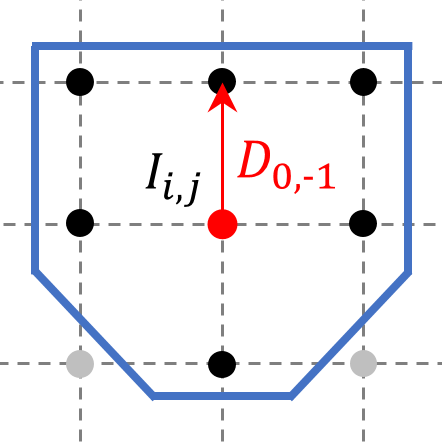}
    \end{minipage}
    \begin{minipage}{0.49\linewidth}
    \includegraphics[width=1\textwidth]{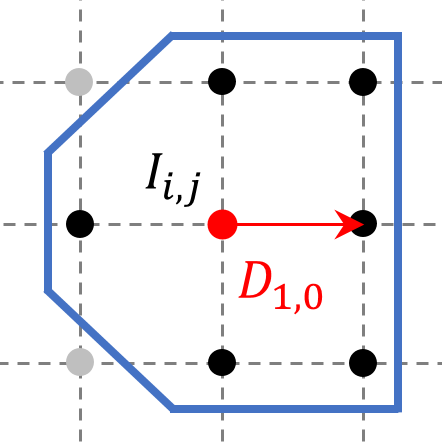}
    \end{minipage}
    \end{minipage}
    
    %%%%%%%%%%%%%%%%%%% ROW 2 %%%%%%%%%%%%%%%%%%%%
    \begin{minipage}{1\linewidth}
    \begin{minipage}{0.31\linewidth}
    \subcaption{ {A kernel \color{blue}{${\mathbf{K}^H}($}{\color{red}{$D_{p,q}$}}{\color{blue}{$)$}} $\subset N_9[I_{i,j}]$} (square grid).}
    \label{fig3a}
    \end{minipage}
    \begin{minipage}{0.02\linewidth}
    \crule[white]{5pt}{1pt}
    \end{minipage}
    \begin{minipage}{0.31\linewidth}
    \subcaption{The kernel {\color{blue}{${\mathbf{K}^H}($}}{\color{red}{$D_{p,q}$}}{\color{blue}{$)$}} (hexagonal grid).}
    \label{fig3b}
    \end{minipage}
    \begin{minipage}{0.02\linewidth}
    \crule[white]{5pt}{1pt}
    \end{minipage}
    \begin{minipage}{0.31\linewidth}
    \subcaption{  {\color{blue}{${\mathbf{K}^H}($}}{\color{red}{$D_{p,q}$}}{\color{blue}{$)$}} with {\color{red}{$D_{p,q} \in \{(-1,0),(1,0),(0,-1),(0,1)\}$}}}\label{fig3c}
    \end{minipage}
    \end{minipage}
\caption{(a) Our proposed kernel. (b) It can approximate a hexagonal kernel by shifting through direction $D$. (c) A set of kernel candidates which are denoted as design pattens ``U",``R",  ``D", ``L" from left to right.}
\vspace{-0.15in}
\label{fig3}
\end{figure}

\begin{figure}[t]
\centering
    \begin{minipage}{0.495\linewidth}
    \includegraphics[width=1\textwidth]{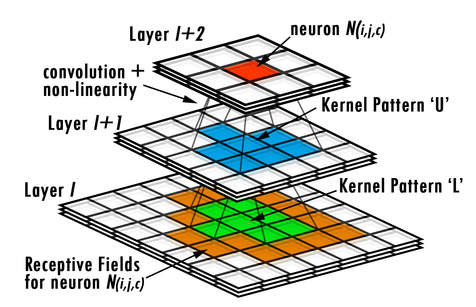}
    \subcaption{}\label{fig4a}
    \vspace{-0.05in}
    \end{minipage}
    \begin{minipage}{0.495\linewidth}
    \includegraphics[width=1\textwidth]{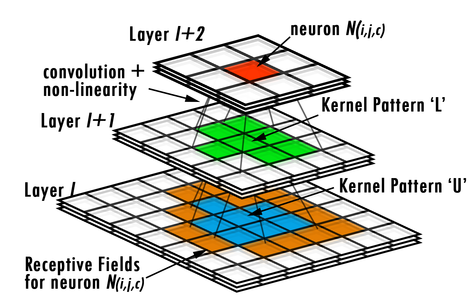}
    \subcaption{}\label{fig4b}
    \vspace{-0.05in}
    \end{minipage}
    
\caption{(a) Employment of the proposed method in CNNs by stacking small size ``quasi-hexagonal" kernels. (b) The kernels employed at different layers of a two-layer CNN will induce the same pattern of RFs on images observed in (a), if only the kernels designed with the same patterns are used, independent of order of their employment.}
\label{fig4}
\vspace{-0.1in}
\end{figure}

\subsection{Designing shape of convolution kernels}
In this work, we address the aforementioned problems by designing shapes of kernels on a two-dimensional coordinate system. For each channel of a given image $\mathbf{I}$, we associate each pixel $I_{i,j} \in \mathbf{I}$ at each coordinate $(i,j)$ with a lattice point (i.e., a point with integer coordinates) in a square grid (Fig.~\ref{fig3a}) \cite{gip,topo}. If two lattice points in the grid are distinct and each $(i,j)$ differs from the corresponding coordinate of the other by at most 1, then they are called 8-adjacent \cite{gip,topo}. An 8-neighbor of a lattice point $I_{i,j} \in \mathbf{I}$ is a point that is 8-adjacent to $I_{i,j}$. We define $N_9[I_{i,j}]$ as a set consisting of a pixel $I_{i,j} \in \mathbf{I}$, and its 8 nearest neighbors (Fig.~\ref{fig3a}). A shape of a \textit{quasi-hexagonal} kernel $\mathbf{K}^H(D_{p,q}) \subset N_9[I_{i,j}]$ is defined as
\begin{equation}
    \mathbf{K}^H(D_{p,q}) = \{I_{i+p,j+q}: I_{i,j} \in N_9[I_{i,j}]\},
\end{equation}
where ${D_{p,q}\in\mathcal{D} }$ is a random variable used as an indicator function employed for designing of shape of $\mathbf{K}^H(D_{p,q})$, and takes values from $\mathcal{D}=\{(-1,0),(1,0),(0,-1),(0,1)\}$ (see Fig.~\ref{fig3c}). Then, convolution of the proposed quasi-hexagonal kernel $\mathbf{K}^H(D_{p,q})$ on a neighborhood centered at a pixel located at $(x,y)$ on an image $\mathbf{I}$ is defined as
\begin{equation}
    I_{x,y}*\mathbf{K}^H(D_{p,q})= \sum_{s,t} \mathbf{K}^H_{s,t}(D_{p,q}) I_{x-s,y-t}.
\end{equation}

%----------------------------------------------------------------------

\subsection{Properties of receptive fields and quasi-hexagonal kernels}
\label{sec:analysis}
Aiming at more flexible representation of shapes of natural objects which may diverge from a fixed square shape, we stack ``quasi-hexagonal'' kernels designed with different shapes, as shown in Fig.~\ref{fig4}. For each convolution layers, we randomly select $D_{p,q}\in \mathcal{D} $ according to a uniform distribution to design kernels. 
%by which a shape that diverge from a fixed square shape, 
%Then, we use the kernels with different orientations at different convolution layers for learning of representations. 
%In order to represent \textit{complex} shapes of natural objects which may diverge from a fixed square shape, we stack ``quasi-hexagonal'' kernels designed with different shapes, to build more versatile ROI using neurons at higher layers of CNNs, as shown in Fig.~\ref{fig4}. For the equipment of learned representations with statistical variability, we randomly select $D_{p,q}\in \mathcal{D} $ according to a uniform distribution to design kernels. Then, we use the kernels at different convolution layers for learning of representations. 
Random selection of design patterns of kernels is feasible because the shapes of RFs will not change, independent of the order of employment of kernels if only the kernels designed with the same patterns are used by the corresponding units (see Fig.~\ref{fig4b}). Therefore, if a CNN model is deep enough, then RFs with a more stable shape will be induced at the last layer, compared to the RFs of middle layer units. 

\begin{figure}[tH]
	\begin{minipage}{1\linewidth}
	\centering
    	\begin{minipage}{0.24\linewidth}
    		\centering
    		\includegraphics[width=0.48\textwidth]{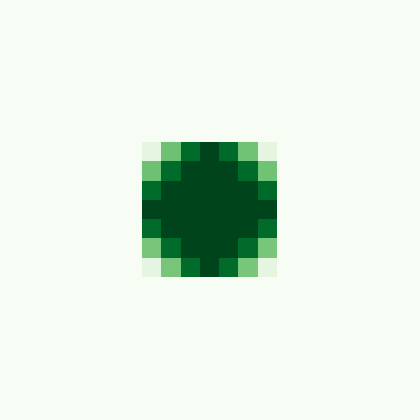}
    		\includegraphics[width=0.48\textwidth]{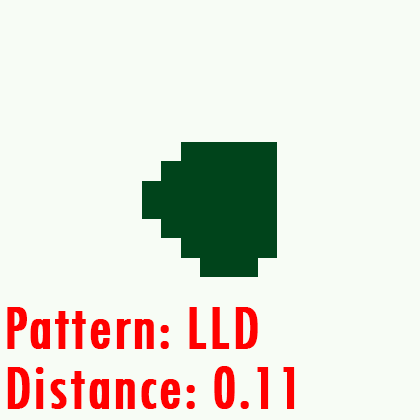}
    % 		\subcaption{$Depth=3$,\\
    % 			$\mathbb{E}[d]=0.075$,\\
    % 			$~\mathbb{V}[d]=0.0014$.}
    		% \mathbb{E}[d]=\scinum{0.075}\\
    		% \mathbb{V}[d]=\scinum{0.0014}$}
    		\label{fig6a}
    	\end{minipage}
    	\begin{minipage}{0.24\linewidth}
    		\centering
    		\includegraphics[width=0.48\textwidth]{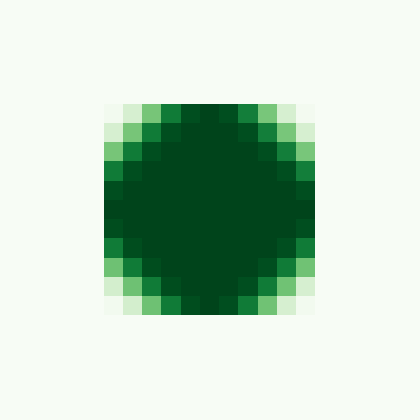}
    		\includegraphics[width=0.48\textwidth]{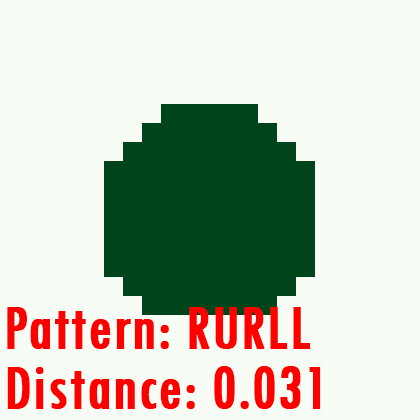}
    % 		\subcaption{$Depth=5$,\\
    % 			$\mathbb{E}[d]=0.061$,\\
    % 			$~\mathbb{V}[d]=0.00092$.}
    		% \mathbb{E}[d]=\scinum{0.061}\\
    		% \mathbb{V}[d]=\scinum{0.00092}$}
    		\label{fig6b}
    	\end{minipage}
    	\begin{minipage}{0.24\linewidth}
    		\centering
    		\includegraphics[width=0.48\textwidth]{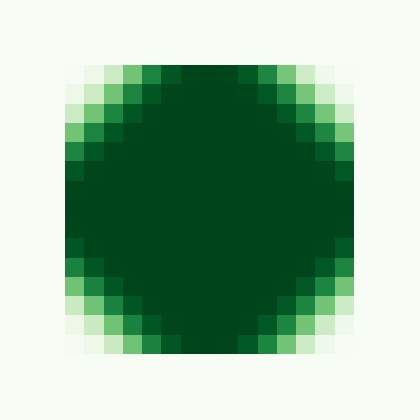}
    		\includegraphics[width=0.48\textwidth]{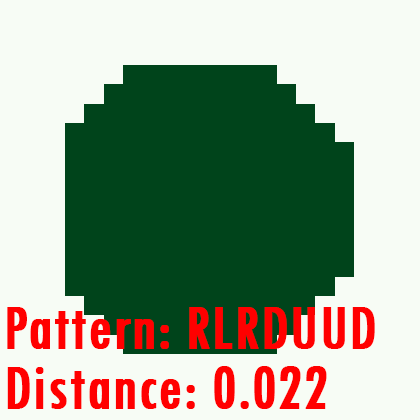}
    % 		\subcaption{$Depth=7$,\\
    % 			$\mathbb{E}[d]=0.053$,\\
    % 			$~\mathbb{V}[d]=0.00069$.}
    		% \mathbb{E}[d]=\scinum{0.053}\\
    		% \mathbb{V}[d]=\scinum{0.00069}$}
    		\label{fig6c}
    	\end{minipage}
    	\begin{minipage}{0.24\linewidth}
    		\centering
    		\includegraphics[width=0.48\textwidth]{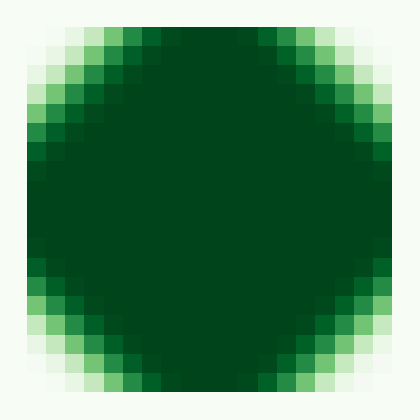}
    		\includegraphics[width=0.48\textwidth]{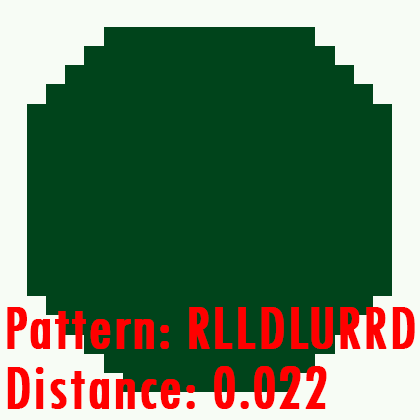}
    % 		\subcaption{$Depth=9$,\\
    % 			$\mathbb{E}[d]=0.046$,\\
    % 			$~\mathbb{V}[d]=0.00054$.}
    		% \mathbb{E}[d]=\scinum{0.046}\\
    		% \mathbb{V}[d]=\scinum{0.00054}$}
    		\label{fig6d}
    	\end{minipage}
    \end{minipage}
	
% 	\begingroup
% 	\captionsetup[subfigure]{font=footnotesize}
	\begin{minipage}{1\linewidth}
	\centering
	    \begin{minipage}{0.24\linewidth}
	    \centering
		\subcaption{$Depth=3$,\\
			$\mathbb{E}[d]=0.075$,\\
			$~\mathbb{V}[d]=0.0014$.}
	    \end{minipage}
	    \begin{minipage}{0.24\linewidth}
	    \centering
		\subcaption{$Depth=5$,\\
			$\mathbb{E}[d]=0.061$,\\
			$~\mathbb{V}[d]=0.00092$.}
	    \end{minipage}
	    \begin{minipage}{0.24\linewidth}
	    \centering
		\subcaption{$Depth=7$,\\
			$\mathbb{E}[d]=0.053$,\\
			$~\mathbb{V}[d]=0.00069$.}
	    \end{minipage}
	    \begin{minipage}{0.24\linewidth}
	    \centering
		\subcaption{$Depth=9$,\\
			$\mathbb{E}[d]=0.046$,\\
			$~\mathbb{V}[d]=0.00054$.}
	    \end{minipage}
	\end{minipage}
	\vspace{-0.05in}
% 	\endgroup
	
\caption{In (a), (b), (c) and (d), the figures given in left and right show an average shape of kernels emerged from 5000 different shape configurations, and a shape of a kernel designed using a single shape configuration, respectively. It can be seen that  the average and variance of $d$ decreases as the kernels are computed at deeper layers. In other words, at deeper layers of CNNs, randomly generated configurations of shapes of kernels can provide better approximations to average shapes of kernels.}
\vspace{-0.15in}
\label{fig5}
\end{figure}

We carry out a Monte Carlo simulation to examine this property using different kernel arrangements. Given an image $\mathbf{I}\in{\mathbb{R}^{W\times H}}$, we first define a stochastic matrix ${\mathcal{M}\in{\mathbb{R}^{W\times H}}}$. The elements of the matrix are random variables ${\mathcal{M}_{i,j}\in{[0,1]}}$ whose values represent the probability that a pixel ${I_{i,j} \in  \mathbf{I}}$ is \textit{covered} by an RF. Next, we define ${\hat{\mathcal{M}} \triangleq \sum_{k} \mathcal{M}_S^k}$ as an average of RFs for a set of kernel arrangements $ \{ \mathcal{M}_S^k \} _{k=1} ^K$. Then, the difference between  $\mathcal{M}_S^k$ and the average $\hat{\mathcal{M}}$ is computed using
\begin{equation}
d(\hat{\mathcal{M}},\mathcal{M}_{S}^k) = \| \hat{\mathcal{M}} - \mathcal{M}_{S}^k \|_F^2/(WH),
\end{equation}
where $\|\cdot \| _F^2$ is the squared Frobenius norm \cite{hartley}. Note that, we obtain a better approximation to the average RF as the distance decreases. The results are depicted in Fig.~\ref{fig5}. The average $\mathbb{E}[d]$ and variance $\mathbb{V}[d]$ show that a better approximation to the average RF is obtained, if kernels used at different layers are integrated at higher depth. 

%----------------------------------------------------------------------
\section{Visualization of regions of interest}
\label{sec:vis}
We propose a method to visualize the features detected in RFs and the ROI of the image. Following the feature visualization approach suggested in \cite{deepinside}, our proposed method provides a saliency map by back-propagating the classification score for a given image and a class. Given a CNN consisting of $L$ layers, the score vector for an input image $\mathbf{I} \in \mathbb{R}^{H \times W \times C}$ is defined as 
\begin{equation}
    \mathbf{S} = F_{1}(\mathbf{W}^{1}, F_{2}(\mathbf{W}^{2},\ldots, F_{L}(I,\mathbf{W}^{L}))),
\end{equation}
where $\mathbf{W}^{L}$ is the weight vector of the kernel $\mathbf{K}_{L}$ at the $L^{th}$ layer, and $S^\mathcal{C}$ is the $\mathcal{C}^{th}$ element of $\mathbf{S}$ representing the classification score for the $\mathcal{C}^{th}$ class. At the $l^{th}$ layer, we compute a feature map $\mathbf{M}^{l}$ for each unit $u^{l}_{i,j,k}\in{\mathbf{M}^{l}}$, which takes values from its receptive field $\mathcal{R}(u^{l}_{i,j,k})$, and  generate a new feature map $\mathbf{\hat M}^{l}$ in which all the units except $u^{l}_{i,j,k}$ are set to be $0$. Then, we feed $\mathbf{\hat M}^{l}$ to the \textit{tail} of the CNN to calculate its score vector as
\begin{equation}
%\begin{split}
    \mathbf{S}(u^{l}_{i,j,k}) = F_{{l+1}}(\mathbf{W}^{{l+1}}, F_{{l+2}}(\mathbf{W}^{{l+2}}, \ldots, F_{L}(\mathbf{\hat M}^{l},\mathbf{W}^{L}))).
%\end{split}
\end{equation}
Thereby, we obtain a score map $\mathbb{S}^{l}$ for all the units of $\mathbf{M}^{l}$, from which we choose top $N$ most contributed units, i.e. the units with the $N$-highest scores. Then, we back-propagate their score $\mathbf{S}^{\mathcal{C}}(u^{l}_{i,j,k})$ for the correct (target) class label  towards the forepart of the CNN to rank the contribution of each pixel $p \in \mathbf{I}$ to the score as
\begin{equation}
%\begin{split}
    \mathbb{S}^{l} (\mathcal{C},u^{l}_{i,j,k}) = F^{-1}_{1}(\mathbf{W}^{1}, F^{-1}_{2}(\mathbf{W}^{{2}}, \ldots, F^{-1}_{l}(\mathbf{S}^{\mathcal{C}}(u^{l}_{i,j,k}),\mathbf{W}^{l}))),
%\end{split}
\end{equation}
where $\mathbb{S}^{l} (\mathcal{C},u^{l}_{i,j,k})$ is a score map that has the same dimension with the image $\mathbf{I}$, and that records the contribution of each pixel $p \in \mathbf{I}$ to the $\mathcal{C}^th$ class. Here we choose the top $\Omega$ unit $\{\mathfrak{u}^l_\omega\}^\Omega_{\omega=1}$ with the highest score $\mathbf{S}^{\mathcal{C}}$, where $\mathfrak{u}^l_\omega$ is the $\omega^{th}$ unit employed at the $l^{th}$ layer. Then, we compute the incorporated saliency map $\mathbf{L}^{\mathcal{C},l}\in\mathbb{R}^{H\times W}$extracted at the $l^{th}$ layer, for the $\mathcal{C}^{th}$ class as follows 
\begin{equation}
    \mathbf{L}^{\mathcal{C},l} = \sum_{\omega} |\mathbb{S}^{l} (\mathcal{C},\mathfrak{u}^{l}_{\omega})|,
\end{equation}
where $| \cdot|$ is the absolute value function. Finally, the ROI of defined by a set of merged RFs, $\{\mathcal{R}(\mathfrak{u}^l_\omega)\}^\Omega_{\omega=1}$ is depicted as a non-zero region in $\mathbf{L}^{\mathcal{C},l}$.

\section{Experiments}
\label{sec:exp}
In Sect.~\ref{sec:perf}, we examine classification performance of CNNs implenmenting proposed methods using two benchmark datasets, CIFAR-10/100\cite{cifar} and ILSVRC-2012 (a subset of ImageNet~\cite{ILSVRC15}). We first analyze the relationship between shape of kernels, ROI and localization of feature detections on images. Then, we  examine the robustness of CNNs for classification of occluded images. Implementation details of the algorithms, and additional results are provided in the supplemental material. We implemented CNN models using the Caffe framework~\cite{caffe}, the QH-conv. layer is implemented by utilizing the im2col method to vectorize the inputs and multiply them with corresponding weight matrix.
\vspace{-0.05in}

%-------------------
% \begin{figure*}[!t]
% \centering
%     \begin{minipage}{1\linewidth}
%     \includegraphics[width=1\textwidth]{eccv/SQU_MINO}
%     \end{minipage}
% \caption{This histogram shows the comparison of class-wise accuracy between BASE and QH-BASE. Each line in the horizontal axis represents accuracy of a class, where a gray segment depicts the top-5 accuracy for base model, the red and green segment depicts an increase and decrease of accuracy, respectively. The proposed model boosts classification accuracy for 52.8\% of the classes, while 22.7\% of the classes shows a decrement in accuracy compared to base model. The top 3 classes with most increment/decrement are mousetrap, lens cap, spotlight and rock crab, conch, fire screen, respectively.}
% \label{fig6}
% \end{figure*}
%-------------------
\subsection{Classification performance}
\label{sec:perf}

\subsubsection{Experiments on CIFAR datasets}
\label{sssec:cifar}

A list of CNN models used in experiments is given in Table~\ref{table1a}. We used the ConvPool-CNN-C model proposed in \cite{Allconv} as our base model (BASE-A). We employed our method in three different models: i) QH-A retains the structure of the BASE-A by just implementing kernels using the proposed methods, ii) QH-B models a larger number of feature maps compared to QH-A such that QH-B and BASE-A have the same number of parameters, iii) QH-C is a larger model which is used for examination of generalization properties (over/under-fitting) of the proposed QH-models. Following~\cite{Allconv} we implement dropout on the input image and at each max pooling layer. We also utilized most of the hyper-parameters suggested in \cite{Allconv} for training the models. We decreased weight decay during the last 100 training epochs to avoid local optima. 

%------------------------------------------------------------------------
\begin{table}[!tH] %% TABLE 1 %%
  \centering
  \caption{CNN configurations. The convolution layer parameters are denoted as \textless Number of duplication\textgreater $\times$conv\textless kernel\textgreater -\textless number of channels\textgreater. A rectified linear unit (ReLU) is followed after each convolution layer. ReLU activation and dropout layer are not shown for brevity. All the conv-3x3/QH/FK layers are set to be stride 1 equipped with pad 1}
  
    \begin{minipage}{1\linewidth}
    \centering
    \subcaption{CNN Configurations - CIFAR}
    \setlength{\tabcolsep}{6pt}
    \label{table1a}
    \begin{tabular}{*1{>{\centering\arraybackslash}c|>{\centering\arraybackslash}c|>{\centering\arraybackslash}c} @{}m{0pt}@{}}
    \noalign{\hrule height 1pt}
    %\multicolumn{3}{c}{CNN Configuration - CIFAR} \\\hline
    \textbf{BASE/BASE-F} &  \textbf{QH-A} & \textbf{QH-B/C}\\\hline
    $3\times$conv-3$\times$3/FK-96 & $3\times$conv-QH-96 & $3\times$convH-108/128 \\\hline
    \multicolumn{3}{c}{maxpool} \\\hline
    $3\times$conv-3$\times$3/FK-192 & $3\times$conv-QH-192 & $3\times$convH-217/256 \\\hline
    \multicolumn{3}{c}{maxpool} \\\hline
    conv-3$\times$3/FK-192 & convH-192 & conv-QH-217/384 \\
    conv-1$\times$1-192 & conv-1$\times$1-192 & conv-1$\times$1-217/384 \\\hline
    \multicolumn{3}{c}{conv1-10/100} \\
    \multicolumn{3}{c}{global avepool + soft-max classifier} \\%\hline
    %\multicolumn{3}{c}{soft-max} \\
    \noalign{\hrule height 1pt}
    \end{tabular}
    \end{minipage}

    \begin{minipage}{1\linewidth}
    \centering
    \subcaption{CNN Configurations - ImageNet}
    \setlength{\tabcolsep}{6pt}
    \label{table1b}
    \begin{tabular}{*1{>{\centering\arraybackslash}c|>{\centering\arraybackslash}c|>{\centering\arraybackslash}c} @{}m{0pt}@{}}
    \noalign{\hrule height 1pt}
    %\multicolumn{3}{c}{CNN Configuration - ImageNet} \\\hline
    \textbf{BASE} & \textbf{QH-BASE} & \textbf{REF-A/B-BASE} \\\hline
    $2\times$conv-3$\times$3-96 & $2\times$conv-QH-96 & $2\times$conv-UB/DIA-96 \\\hline
    \multicolumn{3}{>{\centering\arraybackslash}c}{maxpool} \\\hline
    $2\times$conv-3$\times$3-192 & $2\times$conv-QH-192 & $2\times$conv-UB/DIA-192  \\\hline
    \multicolumn{3}{c}{maxpool} \\\hline
    $2\times$conv-3$\times$3-384 & $2\times$conv-QH-384 & $2\times$conv-UB/DIA-384 \\\hline
    \multicolumn{3}{c}{maxpool} \\\hline
    $2\times$conv-3$\times$3-768 & $2\times$conv-QH-768 & $2\times$conv-UB/DIA-768 \\\hline
    \multicolumn{3}{c}{maxpool} \\\hline
    $2\times$conv-3$\times$3-1536 & $2\times$conv-QH-1536 & $2\times$conv-UB/DIA-1536 \\\hline
    \multicolumn{3}{c}{maxpool} \\\hline
    \multicolumn{3}{c}{conv-3$\times$3-1000} \\
    \multicolumn{3}{c}{conv-1$\times$1-1000} \\
    \multicolumn{3}{c}{global avepool + soft-max classifier} \\%\hline
    %\multicolumn{3}{c}{soft-max} \\
    \noalign{\hrule height 1pt}
    \end{tabular}
    \end{minipage}

  \label{table1}%
\end{table}%

Since our proposed kernels have fewer parameters compared to 3$\times$3 square shaped kernels, by retaining the same structure as BASE-A, QH-A may benefit from the regularization effects brought by less numbers of total parameters that prevent over-fitting. In order to analyze this regularization property of the proposed method, we implemented a reference model, called BASE-REF with conv-FK (fragmented kernel) layer, which has $3\times3$ convolution kernels, and the values of two randomly selected parameters are set to 0 (to keep the number of effective parameters same with quasi-hexagonal kernels). In another reference model (QH-EXT), shape patterns of kernels (Sect.~\ref{sec:method}) are chosen to be the same ($<R, \ldots, R>$ in this implementation). Moreover, we introduced two additional variants of models using i) different kernel sizes for max pooling (-pool4), and ii) an additional dropout layer before global average pooling (-AD). 

\begin{table}[!tH] %% TABLE 2 %%
  \centering
  \caption{Comparison of classification errors using CIFAR-10 dataset (Single models trained without data augmentation)}
    \renewcommand{\arraystretch}{1}
    \begin{tabular}{cccc}
    \toprule
    \textbf{Model} & \specialcell{\textbf{Testing}\\\textbf{Error(\%)}} &
    \textbf{Model} & \specialcell{\textbf{Testing}\\\textbf{Error(\%)}} \\
    \midrule
    BASE-A & \textbf{9.02} & BASE-A-pool4 & \textbf{8.87} \\
    QH-A & 9.10 & QH-A-pool4 & 9.00\\
    BASE-REF  & 9.89 & BASE-A-AD & \textbf{8.71}\\
    QH-EXT & 9.40 & QH-A-AD & 8.79\\
    \bottomrule
    \end{tabular}%
  \label{table2}%
\end{table}%

\begin{table}[!tH] %% TABLE 3 %%
  \centering
  \caption{Comparison of classification error of models using CIFAR-10/100 datasets (Single models trained without data augmentation)}
    \begin{tabular}{cccc}
    \toprule
    \multirow{2}[4]{*}{\textbf{Model}} & \multicolumn{2}{c}{\textbf{Testing Error (\%)}} &  \textbf{Numbers of} \\
          & CIFAR-10 & CIFAR-100 & \textbf{Params.} \\
    \midrule
    NIN~\cite{NIN} & 10.41 & 35.68 & $\approx 1M$\\
    DSN~\cite{DSN} & 9.69  & 34.57 & $\approx 1M$\\
    ALL-CNN~\cite{Allconv} & 9.08  & 33.71 & $\approx 1.4M$\\
    RCNN~\cite{RCNN} & 8.69  & 31.75 & $\approx 1.9M$\\
    Spectral pool~\cite{SpectralPooling} & 8.6 & 31.6 & $-$ \\
    FMP~\cite{FMP} & $-$ & 31.2 & $\approx 12M$ \\
    BASE-A-AD & 8.71 & 31.2 & $\approx 1.4M$ \\
    \textbf{QH-B-AD} & 8.54  & 30.54 & $\approx 1.4M$ \\
    \textbf{QH-C-AD} & \textbf{8.42}  & \textbf{29.77} & $\approx 2.4M$ \\
    \bottomrule
    \end{tabular}
  \label{table3}
\end{table}

Results given in Table~\ref{table2} show that the proposed QH-A has comparable performance to the base CNN models that employ square shape kernels, despite a smaller number of parameters. Meanwhile, a significant decrement in accuracy appears in the BASE-REF model that employs the same number of parameters as QH-A, which suggests that our proposed model works not only by the employment of a regularization effect but by the utilization of a smaller number of parameters. The inferior performance for QH-EXT model indicates the effectiveness of randomly selecting kernels described in Sect.~\ref{sec:method}. Moreover, it can also be observed that the implementation of additional dropout and larger size pooling method improves the classification performance of both BASE-A and proposed QH-A in a similar magnitude. Then, the experimental observation implies a general compatibility between the square kernels and the proposed kernels.

Additionally, we compare the proposed methods with state-of-the-art methods for CIFAR-10 and CIFAR-100 datasets. For CIFAR-100, we used the same models implemented for CIFAR-10 with the same hyper-parameters. The results given in Table~\ref{table4} show that our base model with an additional dropout (BASE-A-AD) provides comparable classification performance for CIFAR-10, and outperforms the state-of-the-art models for CIFAR-100. Moreover, our proposed models (QH-B-AD and QH-C-AD) improve the classification accuracy by adopting more feature maps.

\subsubsection{Experiments on ImageNet}
\label{sssec:imagenet}

We use an up-scale model of BASE-A model for CIFAR-10/100 as our base model, which stacks 11 convolution layers with kernels that have regular 3$\times$3 square shape, that are followed by a 1$\times$1 convolution layer and a global average pooling layer. Then, we modified the base model with three different types of kernels: i) our proposed quasi-hexagonal kernels (denoted as conv-QH layer), ii) reference kernels where we remove an element located at a corner and one of its adjacent elements located at edge of a standard 3$\times$3 square shape kernel (conv-UB), iii) reference kernels where we remove an element from a corner and an element from a diagonal corner of a standard 3$\times$3 square shape kernel (conv-DIA). Notice that unlike the fragmented kernels we employed in the last experiment, these two reference kernels can also be used to generate aforementioned shapes of RFs. However, unlike the proposed quasi-hexagonal kernels, we cannot assure that these kernels can be used to simulate hexagonal processing. Configurations of the CNN models are given in Table \ref{table1a}. %Units employed at the last layer of this kind of deep model have larger RFs can cover a substantial part of an image with less parameters, compared to the lower layer units ~\cite{NIN,Allconv}. 
Dropout~\cite{dropout} is used on an input image at the first layer (with dropout ratio 0.2), and after the last conv-3$\times$3 layer. We employ a simple method for fixing the size of train and test samples to $256 \times 256$~\cite{VGG}, and a patch of $224 \times 224$ is cropped and fed into network during training. Additional data augmentation methods, such as random color shift~\cite{Alexnet}, are not employed for fast convergence. 

\begin{figure}[!tH]
\centering
    %for row 1
    \begin{minipage}[c]{.32\linewidth}
    \centering
    \includegraphics[width=1\textwidth]{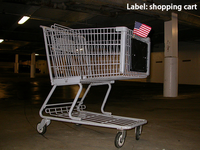}
    \end{minipage}
    \begin{minipage}[c]{.32\linewidth}
    \centering
    \includegraphics[width=1\textwidth]{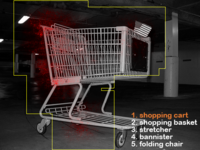}
    \end{minipage}
    \begin{minipage}[c]{.32\linewidth}
    \centering
    \includegraphics[width=1\textwidth]{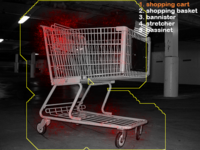}
    \end{minipage}
    
    %for row 2
    \begin{minipage}[c]{.32\linewidth}
    \centering
    \includegraphics[width=1\textwidth]{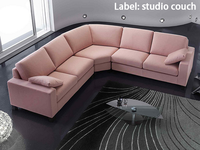}
    \end{minipage}
    \begin{minipage}[c]{.32\linewidth}
    \centering
    \includegraphics[width=1\textwidth]{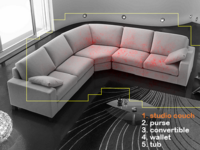}
    \end{minipage}
    \begin{minipage}[c]{.32\linewidth}
    \centering
    \includegraphics[width=1\textwidth]{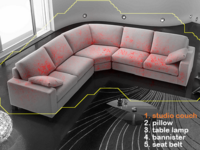}
    \end{minipage}
    
    %for row 3
    \begin{minipage}[c]{.32\linewidth}
    \centering
    \includegraphics[width=1\textwidth]{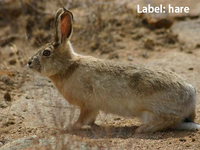}
    \end{minipage}
    \begin{minipage}[c]{.32\linewidth}
    \centering
    \includegraphics[width=1\textwidth]{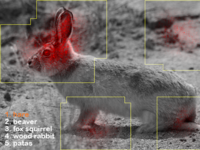}
    \end{minipage}
    \begin{minipage}[c]{.32\linewidth}
    \centering
    \includegraphics[width=1\textwidth]{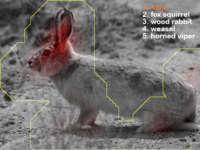}
    \end{minipage}
    
    %for row 4
    \begin{minipage}[c]{.32\linewidth}
    \centering
    \includegraphics[width=1\textwidth]{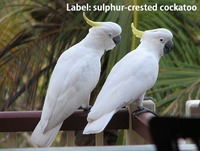}
    \end{minipage}
    \begin{minipage}[c]{.32\linewidth}
    \centering
    \includegraphics[width=1\textwidth]{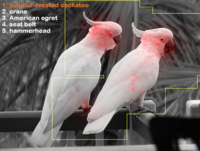}
    \end{minipage}
    \begin{minipage}[c]{.32\linewidth}
    \centering
    \includegraphics[width=1\textwidth]{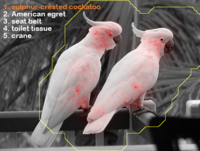}
    \end{minipage}

    %for row 5
    \begin{minipage}[c]{.32\linewidth}
    \centering
    \subcaption*{\textbf{Origin}}
    \end{minipage}
    \begin{minipage}[c]{.32\linewidth}
    \centering
    \subcaption*{\textbf{BASE}}
    \end{minipage}
    \begin{minipage}[c]{.32\linewidth}
    \centering
    \subcaption*{\textbf{QH-BASE}}
    \end{minipage}

\vspace{-0.1in}
\caption{Examples of visualization of ROI. A ROI demonstrates a union of RFs of the top 40 activated neurons at the last max pooling layer. The pixels marked with red color indicate their contribution to classification score, representing the activated features located at them. Borderlines of ROI are represented using yellow frames. Top 5 class predictions provided by the models are also given, and the correct (target) class is given using orange color.}
\label{fig6}
\vspace{-0.18in}
\end{figure}

Classification results are given in Table \ref{table2}. Also, histograms of class-wise accuracy values between BASE and QH-BASE models are given in Fig.~\ref{fig6}. The results show that the performance of reference models is slightly better than that of the base model. Notice that since the base model is relatively over-fitted (top5 accuracy for training sets is $\geq$97\%), these two reference models are more likely to be benefited from the regularization effect brought by less number of parameters. Meanwhile, our proposed QH-BASE outperformed all the reference models, implying the validity of the proposed quasi-hexagonal kernels in approximating hexagonal processing. Detailed analyses concerning compactness of models are provided in the next section.

\begin{table}[!tH] %% TABLE 4 %%
  \centering
  \caption{Comparison of classification performance using validation set of the ILSVRC-2012}
    \begin{tabular}{ccc}
    \toprule
    \textbf{Model} & \textbf{top-1 val.error (\%)} & \textbf{top-5 val.error (\%)} \\
    \midrule
    BASE       & 31.2   & 12.3 \\
    QH-BASE    & \textbf{29.2}   & \textbf{11.1} \\
    REF-A-BASE & 31.4 & 12.4 \\
    REF-B-BASE & 31.2 & 12.2 \\
    \bottomrule
    \end{tabular}%
  \label{table4}%
\end{table}%

\subsubsection{Analysis of relationship between compactness of models and classification performance}
\label{sssec:comp}

In this section, we analyze the compactness of learned models for ImageNet and CIFAR-10 datasets. First, we provide a comparison of the number of parameters and computational time of the models in Table~\ref{table5}. The results show that, in the experimental analyses for the CIFAR-10 dataset, QH-A model has a comparable performance to the base model with fewer parameters and computational time. If we keep the same number of parameters (QH-B), then classification accuracy improves for similar computational time. Meanwhile, in the experimental analyses for the ImageNet dataset, our proposed model shows significant improvement in both model size and computational time.

We conducted another set of experiments to analyze the relationship between the classification performance and the number of training samples using CIFAR-10 dataset. The results given in Table~\ref{table6} show that  the  QH-A-AD model provides a comparable performance with the base model, and the QH-B-AD model provides a better classification accuracy compared to the base model, as the number of training samples decreases. In an extreme case where only 1000 training samples is selected, QH-A-AD and QH-B-AD outperform the base model by 0.7\% and 3.1\%, respectively, which indicates the effectiveness of the proposed method.

\begin{table}[!tH] %% TABLE 5 %%
	\centering
	\caption{Comparison of number of parameters and computational time of different models}
	\begin{tabular}{cccc}
		\toprule
		\textbf{Model} & \specialcell{\textbf{Num. of}\\ \textbf{params.}} &
		\specialcell{\textbf{Training time}\\ \textbf{(500 samples)}} &
		\specialcell{\textbf{Difference}\\ \textbf{in accuracy}}\\
		\midrule
        BASE & $\approx57.3M$ & $51610.5$ ms & $-$\\
		QH-BASE & $\approx44.6M$ & $38815.9$ ms & $+1.2\%$ \\
		\midrule
		BASE-A & $\approx1.4M$ & $1492$ ms & $-$ \\
		QH-A & $\approx1.1M$ & $1227.4$ ms & $-0.08\%$ \\
		QH-B & $\approx1.4M$ & $1449.9$ ms & $+0.17\%$ \\
		\bottomrule
	\end{tabular}
	\label{table5}
\end{table}

\setlength{\tabcolsep}{12pt}
\begin{table}[!tH] %% TABLE 6 %%
\centering
\caption{Comparison of classification error between models BASE-A-AD, QH-A-AD and QH-B-AD with different number of training samples on CIFAR-10 dataset}
\begin{tabular}{cccccc}
\toprule
\multirow{3}[4]{*}{\textbf{Model}} & \multicolumn{5}{c}{\textbf{Classification Error (\%)}} \\
\cmidrule(r){2-6}
& \multicolumn{5}{c}{\textbf{Number of Training Samples}} \\ 
& 20K & 10K & 5K & 2K & 1K \\
\midrule
BASE-AD & 12.6 & 16.8 & 21.8 & 31.0 & 44.9 \\
QH-A-AD & 12.7 & 16.6 & 21.1 & 31.3 & 44.2 \\
QH-B-AD & 12.4 & 16.3 & 20.7 & 30.9 & 41.8 \\
\bottomrule
\end{tabular}
\label{table6}
\end{table}
\setlength{\tabcolsep}{1.4pt}

\begin{figure}[!tH]
\centering
    %for row 1
    \begin{minipage}[c]{.32\linewidth}
    \centering
    \includegraphics[width=1\textwidth]{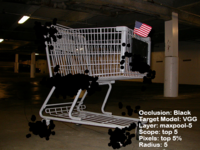}
    \end{minipage}
    \begin{minipage}[c]{.32\linewidth}
    \centering
    \includegraphics[width=1\textwidth]{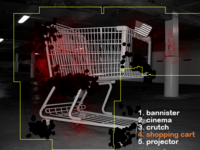}
    \end{minipage}
    \begin{minipage}[c]{.32\linewidth}
    \centering
    \includegraphics[width=1\textwidth]{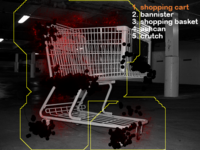}
    \end{minipage}
    
    %for row 2
    \begin{minipage}[c]{.32\linewidth}
    \centering
    \includegraphics[width=1\textwidth]{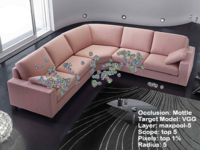}
    \end{minipage}
    \begin{minipage}[c]{.32\linewidth}
    \centering
    \includegraphics[width=1\textwidth]{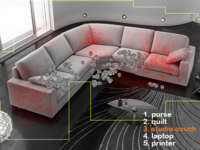}
    \end{minipage}
    \begin{minipage}[c]{.32\linewidth}
    \centering
    \includegraphics[width=1\textwidth]{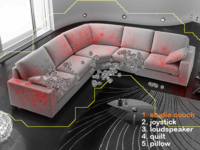}
    \end{minipage}
    
    %for row 3
    \begin{minipage}[c]{.32\linewidth}
    \centering
    \includegraphics[width=1\textwidth]{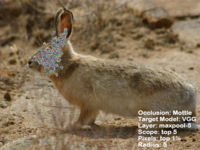}
    \end{minipage}
    \begin{minipage}[c]{.32\linewidth}
    \centering
    \includegraphics[width=1\textwidth]{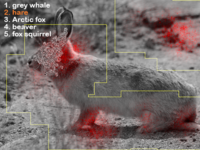}
    \end{minipage}
    \begin{minipage}[c]{.32\linewidth}
    \centering
    \includegraphics[width=1\textwidth]{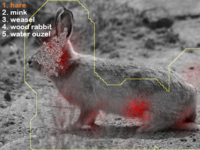}
    \end{minipage}
    
    %for row 4
    \begin{minipage}[c]{.32\linewidth}
    \centering
    \includegraphics[width=1\textwidth]{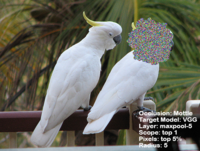}
    \end{minipage}
    \begin{minipage}[c]{.32\linewidth}
    \centering
    \includegraphics[width=1\textwidth]{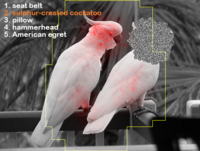}
    \end{minipage}
    \begin{minipage}[c]{.32\linewidth}
    \centering
    \includegraphics[width=1\textwidth]{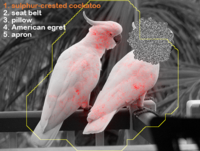}
    \end{minipage}

    %for row 5
    \begin{minipage}[c]{.32\linewidth}
    \centering
    \subcaption*{\textbf{Occluded Image}}
    \end{minipage}
    \begin{minipage}[c]{.32\linewidth}
    \centering
    \subcaption*{\textbf{BASE}}
    \end{minipage}
    \begin{minipage}[c]{.32\linewidth}
    \centering
    \subcaption*{\textbf{QH-BASE}}
    \end{minipage}

\vspace{-0.1in}
\caption{Analysis of robustness of different models to occlusion. We use the same proposed method to select neurons and visualize their RFs for each model (see Sect.~\ref{sec:vis}). The comparison between the ROI shown in Fig.~\ref{fig6} suggests that the proposed model overcomes the occlusion by detecting features that are spatially distributed on target objects. It can also be seen that, the classification accuracy of the base model is decreased although the ROI of the base model seems to be more adaptive to the shape of objects. This also suggests that the involvement of background may make the CNNs hard to discriminate background from useful features.}
\vspace{-0.25in}
\label{fig7}
\end{figure}
%----------------------------------------------
\subsection{Visualization of regions of interest}

Fig.~\ref{fig6} shows some examples of visualizations depicted using our method proposed in Sect.~\ref{sec:vis}. Saliency maps are normalized and image contrast is slightly raised to improve visualization of images. We observed that for most of these \textit{correctly} classified testing images, both the BASE model equipped with square kernels and the proposed QH-BASE model equipped with quasi-hexagonal kernels are able to present an ROI that roughly specify the location and some basic shape of the target objects, and vise versa. Since the ROI is directly determined by RFs of neurons with strong reactions toward special features, this observation suggests that the relevance between learned representations and target objects is crucial for recognition and classification using large-scale datasets of natural images such as ImageNet.

However, some obvious difference between the ROI of the base model and the proposed model can be observed: i) ROI of the base model usually involves more  background than that of the proposed model. That is, compared to these pixels with strong contributions, the percentage of these pixels that are not essentially contributing to the classification score, is generally higher in the base model. ii) Features learned using the square kernels are more like to be detected within clusters on special parts of the objects. The accumulation of the features located in these clusters  results in a superior contribution, compared to the features that are scattered on the images. For instance, in the base model, more neurons have their RFs located in the heads of hare and parrots, thus the heads obtain higher classification scores than other parts of body. iii) As a result of ii), some duplicated important features (e.g, the supporting parts of cart and seats of coach) are overlooked in these top reacted high-level neurons in the base model. Meanwhile, our proposed model with quasi-hexagonal kernels is more likely to obtain \textit{discriminative} features that are spatially distributed on the whole object. In order to further analyze the results obtained by employing the square kernel and the proposed kernels for object recognition, we provide a set of experiments using occluded images in the next section.

\subsection{Occlusion and spatially distributed representations}
The analyses given in the last section imply that the base CNN models equipped with the square kernel could be vulnerable to  recognition of objects in occluded scenes, which is a very common scenario in computer version tasks. In order to analyze the robustness of the methods to partial occlusion of images, we prepare a set of locally occluded images using the following methods. i) We randomly select 1249 images that are correctly classified by both the base and proposed models using the validation set of ILSVRC-2012 \cite{ILSVRC15}. ii) We select Top1 or Top5 elements with highest classification score at the last maxpool layers of a selected model\footnote{In addition to the BASE and the QH-BASE models, we also employ a ``third-party" model, namely VGG~\cite{VGG}, to generate the occluded images.} and calculate the ROI defined by their RFs, as we described in Sect.~\ref{sec:vis}. iii) Within the ROI, we choose 1-10\% of pixels that provide the most contribution, and then occlude each of the selected pixels with a small circular  occlusion mask (with radius $r=5$ pixels), which is filled by black (Bla.) or randomly generated colors (Mot.) drawn from a uniform distribution. In total, we generate 120 different occlusion datasets (149880 different occluded images in total), Table~\ref{table7} shows the classification accuracy on the occluded images. The results show that our proposed quasi-hexagonal kernel model reveal better robustness in this object recognition under targeted occlusion task compared to square kernel model. Some sample images are shown in Fig.~\ref{fig7}.

\begin{table}[!tH] %% TABLE 7 %%
\centering
\caption{Performances on the occlusion datasets. Each column shows the classification accuracy (\%) of test models in different occlusion conditions. In the first row, BASE/QH-BASE/VGG indicate the models used for generating occlusion, Top1/Top5 indicate the numbers of selected neurons that control the size of occluded region, Bla./Mot. indicate the patterns of occlusion}
\begin{tabular}{cccccccccccccc}
\toprule
\multirow{3}[4]{*}{\textbf{Model}} & \multicolumn{4}{c}{\textbf{BASE}} & \multicolumn{4}{c}{\textbf{QH-BASE}} & \multicolumn{4}{c}{\textbf{VGG}} &
\multirow{3}[4]{*}{\specialcell{\textbf{Average}\\ \textbf{accuracy}}} \\
%\multirow{3}[4]{*}{\textbf{Ave. accuracy}} \\
\cmidrule(r){2-5}
\cmidrule(r){6-9}
\cmidrule(r){10-13}
& \multicolumn{2}{c}{Top1} & \multicolumn{2}{c}{Top5}
& \multicolumn{2}{c}{Top1} & \multicolumn{2}{c}{Top5}
& \multicolumn{2}{c}{Top1} & \multicolumn{2}{c}{Top5} & \\
& Bla. & Mot. & Bla. & Mot. & Bla. & Mot. & Bla. & Mot. & Bla. & Mot. & Bla. & Mot. &\\
\midrule
BASE & 58.8 & 61.2 & 34.6 & 40.9 & 61.3 & 63.5 & 36.3 & 42.7 & 61.7 & 63.8 & 44.1 & 48.3 & 51.5\\
QH-BASE & 67.1 & 67.8 & 43.8 & 47.6 & 67.0 & 66.9 & 42.2 & 45.4 & 68.6 & 69.1 & 52.3 & 54.6 & \textbf{57.7}\\
\bottomrule
\end{tabular}
\vspace{-0.1in}
\label{table7}
\end{table}

%------------------------------------------------------------------------
\section{Conclusion}
\label{sec:conc}
In this work, we analyze the effects of shapes of convolution kernels on feature representations learned in CNNs and classification performance. We first propose a method to design the shape of kernels in CNNs. We then propose a feature visualization method for visualization of pixel-wise classification score maps of learned features. It is observed that the compact representations obtained using the proposed kernels are beneficial for the classification accuracy. In the experimental analyses, we obtained state-of-the-art performance using ImageNet and CIFAR datasets. Moreover, our proposed methods enable us to implement CNNs with less number of parameters and computational time compared to the base-line CNN models. Additionally, the proposed method improves the robustness of the base-line models to occlusion for classification of partially occluded images. These results confirm the effectiveness of the proposed method for designing of the shape of convolution kernels in CNNs for image classification. In future work, we plan to apply the proposed method to perform other tasks such as object detection and segmentation.

\clearpage

\appendix
\section{Introduction}
\label{sec:intro}

In this document, we provide the supplemental material for the paper ``Design of Kernels in Convolutional Neural Networks for Image Classification''. In the next section, implementation details of the algorithms proposed and employed in the main text are given. In Section \ref{sec:vis}, additional results for visualization of receptive fields are provided.
%------------------------------------------------------------------------
\section{Implementation detail}
\label{sec:implement}

\subsection{CNN models implemented for CIFAR}

In this subsection, implementation details of the algorithms and models employed in Sect. 4.1.1 of the main text are given.

In a training phase, we optimise a soft-max loss function at the top layer of a CNN model using stochastic gradient descent with mini-batch 128, and a momentum ~\cite{momentum} of 0.9 is used.
All the models are regularized by weight decay (L2 penalty) with multiplier 0.001 initially. For the QH-models, we decrease the multiplier during the final 100 training epochs to avoid local optima. We also regularize all the models using dropout; dropout with ratio 0.2 is employed for input data, and dropout with ratio 0.5 is employed for each maxpool layer.
The learning rate is initially set to $5\times 10^{−2}$, and then decreased by a factor of 10 after 120, 170, and 220 training epochs. The learning algorithm was stopped after 270 epochs with a final learning rate $5\times 10^{-5}$.

We apply the global contrast normalization and ZCA whitening which were implemented by Goodfellow~\etal ~in the maxout network~\cite{Maxout}, and no further data augmentation is employed for both training and testing images.

\subsection{CNN models implemented for Imagenet}

In this subsection, implementation details of the algorithms and models employed in Sect. 4.1.2 of the main text are given.

In a training phase, we optimise a soft-max loss function at the top layer of a CNN model using stochastic gradient descent with mini-batch size 192, and a momentum~\cite{momentum} of 0.9 is used.
In order to regularize these models, we implement weight decay (L2 penalty) and dropout. For VGG and QH-VGG, weight decay multipliers are set to 0.0005, and dropout with ratio 0.5 is employed for the first two fully connected (fc) layers. For QH-GAP, a weight decay multiplier is set to 0.0001,  dropout with ratio 0.5 and 0.2 are employed for the conv3-1000 layer and input data, respectively.
The learning rate is initially set to $10^{−2}$, and then decreased by a factor of 10 when the improvement of a validation set accuracy is stopped. The training was stopped after 65 epochs with a final learning rate $10^{-5}$.

Additionally, in a training phase of a CNN, a training image is first resized to a fixed 256$\times$256, then a 224$\times$224 piece is randomly cropped and mirrored as the CNN receives an input image at each iteration. Further augmentation of training data such as random RGB color shift \cite{Alexnet} is not employed for a fast convergence.
During testing, all the testing images are resized to 256$\times$256, and fed to the CNNs without cropping. Fc layers of VGG and QH-VGG models are converted into convolution layers with kernel size $7\times 7$, hence we could obtain a class score map whose number of channels is equal to the number of classes. Then, the score map is channel-wise average pooled, and fed into the soft-max classifier.
We augment the test images by mirroring, and the final score for a test image is averaged from the original and mirrored images.

\section{Visualization of regions of interest}
\label{sec:vis}

In this subsection, we provide additional results obtained using our proposed visualization method given in Sect. 3 and employed in Sect. 4.2 of the main text.

Additional visualization results of receptive fields are shown in Figure~\ref{fig1}, Figure~\ref{fig2}, Figure~\ref{fig3}.

\begin{figure}[!tH]
\centering
    %for row 1
    \begin{minipage}[c]{.32\linewidth}
    \centering
    \includegraphics[width=1\textwidth]{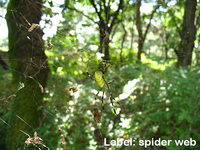}
    \end{minipage}
    \begin{minipage}[c]{.32\linewidth}
    \centering
    \includegraphics[width=1\textwidth]{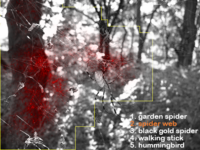}
    \end{minipage}
    \begin{minipage}[c]{.32\linewidth}
    \centering
    \includegraphics[width=1\textwidth]{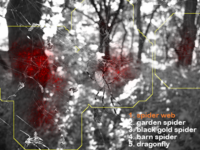}
    \end{minipage}
    \begin{minipage}[c]{1\linewidth}
    \crule[white]{10pt}{1pt}
    \end{minipage}
    
    %for row 2
    \begin{minipage}[c]{.32\linewidth}
    \centering
    \includegraphics[width=1\textwidth]{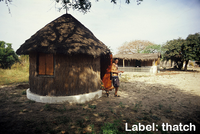}
    \end{minipage}
    \begin{minipage}[c]{.32\linewidth}
    \centering
    \includegraphics[width=1\textwidth]{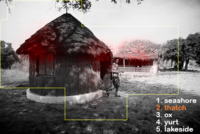}
    \end{minipage}
    \begin{minipage}[c]{.32\linewidth}
    \centering
    \includegraphics[width=1\textwidth]{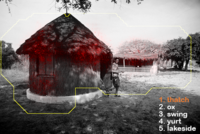}
    \end{minipage}
    \begin{minipage}[c]{1\linewidth}
    \crule[white]{10pt}{1pt}
    \end{minipage}
    
    %for row 3
    \begin{minipage}[c]{.32\linewidth}
    \centering
    \includegraphics[width=1\textwidth]{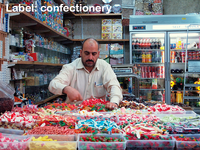}
    \end{minipage}
    \begin{minipage}[c]{.32\linewidth}
    \centering
    \includegraphics[width=1\textwidth]{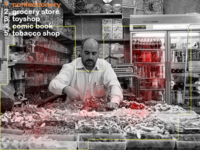}
    \end{minipage}
    \begin{minipage}[c]{.32\linewidth}
    \centering
    \includegraphics[width=1\textwidth]{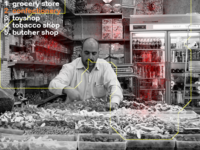}
    \end{minipage}
    \begin{minipage}[c]{1\linewidth}
    \crule[white]{10pt}{1pt}
    \end{minipage}
    
    %for row 4
    \begin{minipage}[c]{.32\linewidth}
    \centering
    \includegraphics[width=1\textwidth]{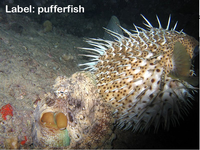}
    \end{minipage}
    \begin{minipage}[c]{.32\linewidth}
    \centering
    \includegraphics[width=1\textwidth]{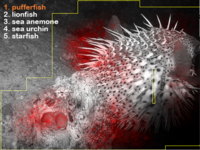}
    \end{minipage}
    \begin{minipage}[c]{.32\linewidth}
    \centering
    \includegraphics[width=1\textwidth]{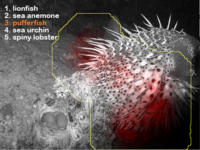}
    \end{minipage}

    %for row 5
    \begin{minipage}[c]{.32\linewidth}
    \centering
    \subcaption*{\textbf{Origin}}
    \end{minipage}
    \begin{minipage}[c]{.32\linewidth}
    \centering
    \subcaption*{\textbf{BASE}}
    \end{minipage}
    \begin{minipage}[c]{.32\linewidth}
    \centering
    \subcaption*{\textbf{QH-BASE}}
    \end{minipage}

\vspace{-0.1in}
\caption{Additional results for visualization of RFs. See Sect. 4.2 and Figure 6 in the main text for details.}
\label{fig1}
\vspace{-0.18in}
\end{figure}

\begin{figure}[!tH]
\centering
    %for row 1
    \begin{minipage}[c]{.32\linewidth}
    \centering
    \includegraphics[width=1\textwidth]{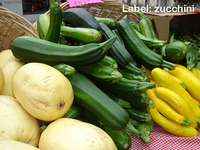}
    \end{minipage}
    \begin{minipage}[c]{.32\linewidth}
    \centering
    \includegraphics[width=1\textwidth]{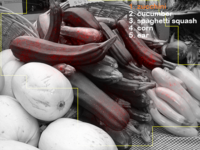}
    \end{minipage}
    \begin{minipage}[c]{.32\linewidth}
    \centering
    \includegraphics[width=1\textwidth]{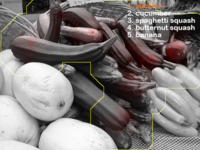}
    \end{minipage}
    \begin{minipage}[c]{1\linewidth}
    \crule[white]{10pt}{1pt}
    \end{minipage}
    
    %for row 2
    \begin{minipage}[c]{.32\linewidth}
    \centering
    \includegraphics[width=1\textwidth]{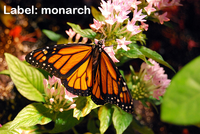}
    \end{minipage}
    \begin{minipage}[c]{.32\linewidth}
    \centering
    \includegraphics[width=1\textwidth]{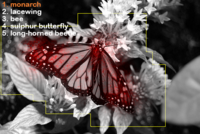}
    \end{minipage}
    \begin{minipage}[c]{.32\linewidth}
    \centering
    \includegraphics[width=1\textwidth]{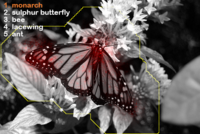}
    \end{minipage}
    \begin{minipage}[c]{1\linewidth}
    \crule[white]{10pt}{1pt}
    \end{minipage}
    
    %for row 3
    \begin{minipage}[c]{.32\linewidth}
    \centering
    \includegraphics[width=1\textwidth]{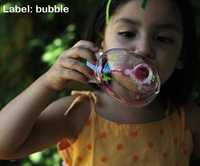}
    \end{minipage}
    \begin{minipage}[c]{.32\linewidth}
    \centering
    \includegraphics[width=1\textwidth]{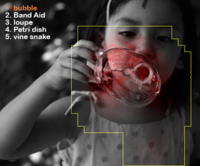}
    \end{minipage}
    \begin{minipage}[c]{.32\linewidth}
    \centering
    \includegraphics[width=1\textwidth]{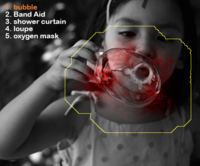}
    \end{minipage}
    \begin{minipage}[c]{1\linewidth}
    \crule[white]{10pt}{1pt}
    \end{minipage}
    
    %for row 4
    \begin{minipage}[c]{.32\linewidth}
    \centering
    \includegraphics[width=1\textwidth]{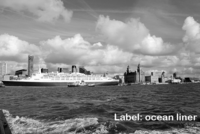}
    \end{minipage}
    \begin{minipage}[c]{.32\linewidth}
    \centering
    \includegraphics[width=1\textwidth]{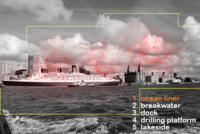}
    \end{minipage}
    \begin{minipage}[c]{.32\linewidth}
    \centering
    \includegraphics[width=1\textwidth]{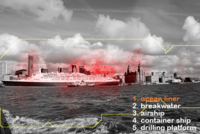}
    \end{minipage}

    %for row 5
    \begin{minipage}[c]{.32\linewidth}
    \centering
    \subcaption*{\textbf{Origin}}
    \end{minipage}
    \begin{minipage}[c]{.32\linewidth}
    \centering
    \subcaption*{\textbf{BASE}}
    \end{minipage}
    \begin{minipage}[c]{.32\linewidth}
    \centering
    \subcaption*{\textbf{QH-BASE}}
    \end{minipage}

\vspace{-0.1in}
\caption{Additional results for visualization of RFs. See Sect. 4.2 and Figure 6 in the main text for details.}
\label{fig2}
\vspace{-0.18in}
\end{figure}

\begin{figure}[!tH]
\centering
    %for row 1
    \begin{minipage}[c]{.32\linewidth}
    \centering
    \includegraphics[width=1\textwidth]{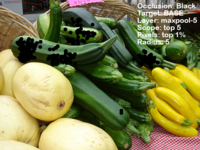}
    \end{minipage}
    \begin{minipage}[c]{.32\linewidth}
    \centering
    \includegraphics[width=1\textwidth]{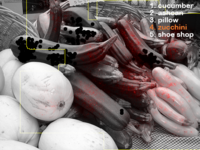}
    \end{minipage}
    \begin{minipage}[c]{.32\linewidth}
    \centering
    \includegraphics[width=1\textwidth]{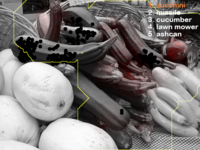}
    \end{minipage}
    \begin{minipage}[c]{1\linewidth}
    \crule[white]{10pt}{1pt}
    \end{minipage}
    
    %for row 2
    \begin{minipage}[c]{.32\linewidth}
    \centering
    \includegraphics[width=1\textwidth]{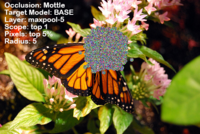}
    \end{minipage}
    \begin{minipage}[c]{.32\linewidth}
    \centering
    \includegraphics[width=1\textwidth]{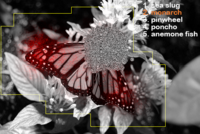}
    \end{minipage}
    \begin{minipage}[c]{.32\linewidth}
    \centering
    \includegraphics[width=1\textwidth]{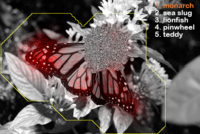}
    \end{minipage}
    \begin{minipage}[c]{1\linewidth}
    \crule[white]{10pt}{1pt}
    \end{minipage}
    
    %for row 3
    \begin{minipage}[c]{.32\linewidth}
    \centering
    \includegraphics[width=1\textwidth]{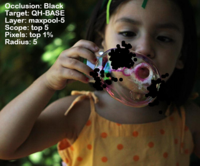}
    \end{minipage}
    \begin{minipage}[c]{.32\linewidth}
    \centering
    \includegraphics[width=1\textwidth]{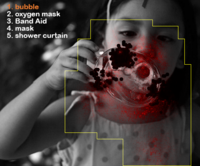}
    \end{minipage}
    \begin{minipage}[c]{.32\linewidth}
    \centering
    \includegraphics[width=1\textwidth]{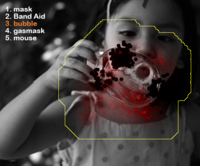}
    \end{minipage}
    \begin{minipage}[c]{1\linewidth}
    \crule[white]{10pt}{1pt}
    \end{minipage}
    
    %for row 4
    \begin{minipage}[c]{.32\linewidth}
    \centering
    \includegraphics[width=1\textwidth]{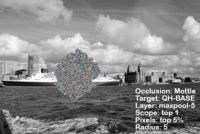}
    \end{minipage}
    \begin{minipage}[c]{.32\linewidth}
    \centering
    \includegraphics[width=1\textwidth]{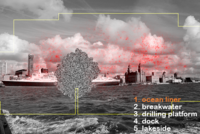}
    \end{minipage}
    \begin{minipage}[c]{.32\linewidth}
    \centering
    \includegraphics[width=1\textwidth]{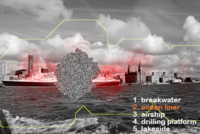}
    \end{minipage}

    %for row 5
    \begin{minipage}[c]{.32\linewidth}
    \centering
    \subcaption*{\textbf{Occluded Image}}
    \end{minipage}
    \begin{minipage}[c]{.32\linewidth}
    \centering
    \subcaption*{\textbf{BASE}}
    \end{minipage}
    \begin{minipage}[c]{.32\linewidth}
    \centering
    \subcaption*{\textbf{QH-BASE}}
    \end{minipage}

\vspace{-0.1in}
\caption{Additional results for visualization of occluded images. See Sect. 4.3 and Figure 7 in the main text for details.}
\label{fig3}
\vspace{-0.18in}
\end{figure}

\clearpage

\bibliographystyle{splncs}
\bibliography{arivx}
\end{document}